\documentclass[lettersize,journal]{IEEEtran}
\usepackage{amsmath,amsfonts}
\usepackage{algorithmicx}
\usepackage{algpseudocode}
\usepackage{algorithm}
\usepackage{array}
\usepackage[caption=false,font=normalsize,labelfont=sf,textfont=sf]{subfig}
\usepackage{textcomp}
\usepackage{stfloats}
\usepackage{url}
\usepackage{verbatim}
\usepackage[dvipsnames]{xcolor}
\usepackage{graphicx}
\usepackage{overpic}
\usepackage{tcolorbox}
\usepackage{booktabs}

\usepackage[
    backend=biber,
    style=ieee,
    citestyle=ieee-comp,
    doi=false,
    url=false,
    isbn=false,
    eprint=false
]{biblatex}

\AtEveryBibitem{%
  \clearfield{issn}%
}

\usepackage{hyperref}
\hypersetup{pdfauthor={}, pdftitle={Chess on Ice}, pdfsubject={}, pdfkeywords={}}

\newcommand{\action}[1]{a_{#1}}
\newcommand{\state}[1]{s_{#1}}

\definecolor{bluecurling}{HTML}{2A6B8F}

\begin{document}

\title{Chess on Ice: Curling Tactical Decision-Making via Backward Induction and Deep Reinforcement Learning}

\author{
Patrick Oberlin,~\IEEEmembership{Student Member,~IEEE,}
Matteo Cederle,~\IEEEmembership{Graduate Student Member,~IEEE,}
Aren Karapetyan,~\IEEEmembership{Member,~IEEE,}
Saverio Bolognani,~\IEEEmembership{Member,~IEEE,}
Gian Antonio Susto,~\IEEEmembership{Senior Member,~IEEE,}
and Florian Dörfler,~\IEEEmembership{Senior Member,~IEEE}
\thanks{Patrick Oberlin, Saverio Bolognani, and Florian Dörfler are with the Automatic Control Laboratory, ETH Zurich, Switzerland.}
\thanks{Aren Karapetyan is with Belimo Automation AG, Hinwil, Switzerland.}
\thanks{Matteo Cederle and Gian Antonio Susto are with the Department of Information Engineering, University of Padova, Italy.}
\thanks{This work was supported by NCCR Automation, grant agreement 51NF40\_225155 from the Swiss National Science Foundation, and by the Ing. Aldo Gini Foundation.}
}

% The paper headers
\markboth{Pre-Print Under Review}%
{Oberlin \MakeLowercase{\textit{et al.}}: Chess on Ice: Curling Tactical Decision-Making via Backward Induction and Deep Reinforcement Learning}

%\IEEEpubid{0000--0000/00\$00.00~\copyright~2021 IEEE}
% Remember, if you use this you must call \IEEEpubidadjcol in the second
% column for its text to clear the IEEEpubid mark.

\maketitle

\begin{abstract}
Curling is often referred to as ``Chess on Ice'', owing to the tactical complexity of its decision-making process. Yet unlike chess, curling remains largely underexplored from a machine learning perspective, with prior work confined mainly to statistical approaches. We propose a reinforcement learning framework capable of quantitatively evaluating and comparing tactical options in curling. The game poses several modeling challenges: continuous state and action spaces, stochastic action outcomes reflecting player skill variability, and state transitions that are highly sensitive to small perturbations in the executed action. To address them, we employ the Deep Deterministic Policy Gradient actor-critic algorithm, adapted to exploit the finite-horizon structure of the game. Our experiments show that effective curling strategies can be acquired in a fully self-supervised manner, without any human-annotated data: on a reduced four-rock variant, the learned agent matches a hand-crafted expert heuristic in a regime where that heuristic is close to optimal, a parity we quantify against the intrinsic hammer advantage of the variant. Beyond the resulting policy, the learned critic provides a dense value estimate over the entire continuous action space, enabling the quantitative comparison of tactical alternatives for applications such as post-game performance analysis and decision support during athlete preparation.

\end{abstract}


\begin{figure}[tb]
    \centering
    \includegraphics[scale=1]{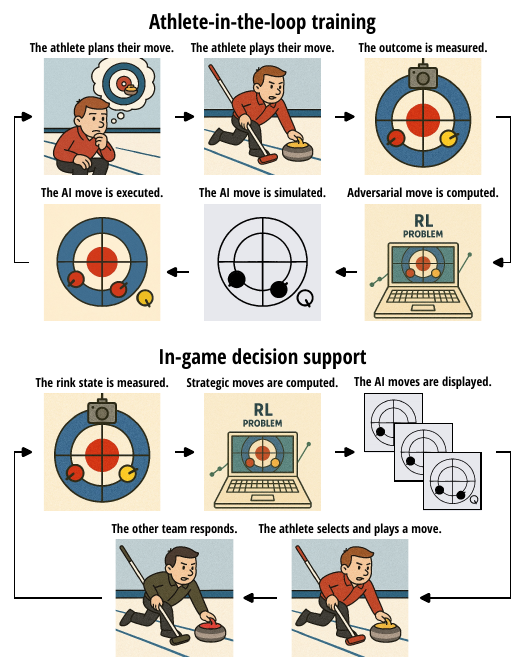}
    \caption{Schematic depiction of possible applications for the proposed tool.}
    \label{fig:athlete-in-the-loop}
\end{figure}

\section{Introduction} \label{sec:introduction}
\IEEEPARstart{G}ames, with their well-defined objectives and reward structures, serve as a natural testbed for reinforcement learning (RL) algorithms \cite{silver2017mastering, mnih2013, li2025comprehensive, wang2025intelligent}. While the history of RL in games dates back to the $1950$-s with the game of checkers \cite{samuel1959some}, and spans several decades with multiple prominent applications, such as backgammon \cite{tesauro1994td}, blackjack~\cite{perez1998blackjack} and chess \cite{campbell2002deep}, it is the AlphaGo algorithm \cite{silver2016mastering} that marked a true breakthrough, achieving remarkable results for the game of Go and chess through a combination of tree-search and Deep Q-Learning methods. While initially restricted to discrete state and action spaces and deterministic transitions due to the nature of the games, these methods have since been extended to continuous settings, with video games \cite{mnih2013, gallotta2024large, li2025comprehensive} taking the center stage. Furthermore, recent advances in large language models, computer vision, robotics, and artificial intelligence (AI) more broadly \cite{vaswani2017attention, naveed2025comprehensive} have further pushed RL to new frontiers, with methods now being applied to physical agents and driving growing interest in physical AI \cite{radanliev2021artificial}. This setting comes with a number of additional challenges that are not present in classical game-theoretic RL formulations with discrete state-action spaces;
examples include learning-based methods for real-world billiards \cite{haar2020motor, schiott2025pix2pockets}, table tennis \cite{tebbe2021sample, dambrosio2025achieving}, badminton \cite{wang2024coachai, ma2025learning}, robot soccer \cite{riedmiller2009reinforcement}, and what is oftentimes referred to as chess on ice: curling. Curling has been approached both from a physical-execution standpoint, e.g., through a robotic system achieving human-level performance against elite athletes \cite{won2020adaptive}, and from a tactical decision-making standpoint \cite{curling:markov, curling:markov2, rl-in-curling}. In this paper, we focus on the latter (whose rules are briefly summarized in the colored box, and available at \cite{curlingrules}) and study how existing RL methods can leverage the particular structure of the game to overcome at least some of its associated challenges and learn tactically meaningful play from self-play alone. Crucially, unlike prior works \cite{curling:markov, curling:markov2, rl-in-curling}, which rely on real gameplay datasets for training, our approach learns entirely from scratch without any pre-collected data.

\begin{figure}[tb]
\begin{tcolorbox}[
    title={The Game of Curling},
    colback=bluecurling!5!white,
    colframe=bluecurling,
    fonttitle=\bfseries,
    rounded corners,
    arc=4pt
]
Curling is a team sport played on ice, where two teams of four players take turns sliding heavy granite \textit{rocks} towards concentric circles, collectively known as the \textit{house}.
The objective is to place one's rocks closer to the center of the house, also called the \textit{button}, than the opponent's best rock. 

A full game of curling typically consists of eight or ten rounds, or \textit{ends}.
During each end, both teams alternately deliver eight rocks to the \textit{house}, resulting in a total of sixteen rocks played.
Only after these sixteen rocks have been played, the score is counted as follows.
The team whose rock is closest to the button scores a point for every rock that is in the house and closer to the button than the best rock of the opposing team.
The other team scores 0 points.
It is possible (sometimes desired) that no rocks remain in the house.
In this case, both teams score 0 points (\textit{blank end}). 

Throwing the last rock is an advantage as it allows to place the rocks favorably right before the score is evaluated.
This right is called \textit{hammer}.
If the team with hammer scores, it loses this right for the following end.
In case of a blank end or if the team without hammer scored, the hammer stays with the same team.
\end{tcolorbox}
\end{figure}

The reason why curling is often compared to chess is its emphasis on planning, positioning, and anticipating the opponent's moves. Crucially, however, unlike chess, in curling the states and actions are continuous, the transition function is non-deterministic, and the game ends after a fixed number of shots, and only then is the score evaluated. This allows tactical planning in the game to unfold over multiple shots, with early rocks deliberately placed as guards. These do not contribute directly to scoring; instead, they protect later scoring attempts by obstructing direct paths for \textit{takeout}.
This layered tactical structure introduces additional depth, as players must reason about multiple future interactions under uncertainty.
The interplay between offensive play (e.g., placing rocks as or behind guards) and defensive responses (e.g., \textit{clearing} guards) requires teams to continuously update their strategy based on the evolving layout of rocks and their risk tolerance.

Reinforcement learning addresses sequential decision-making problems typically formalized as Markov decision processes (MDPs), where the objective is to learn a policy that maximizes the expected cumulative reward \cite{sutton1998reinforcement, puterman2014markov}. Classical approaches, such as dynamic programming \cite{bertsekas2012dynamic} or temporal-difference methods including Q-learning \cite{watkins1992q, sutton1998reinforcement}, establish convergence guarantees in tabular settings but assume discrete state–action spaces with a notable exception of the linear quadratic regulator \cite{bradtke1992reinforcement}, that is a special case of the more general RL problem. To address the  scalability of such methods, function approximations, often with deep neural networks, have been incorporated into value-based methods, such as the Deep Q-Networks (DQN) \cite{mnih2013}, though these remain inherently restricted to discrete actions. Policy gradient methods \cite{sutton1999policy, kakade2001natural, schulman2015trust} overcome this limitation by directly optimizing parameterized policies via gradient ascent in the policy space, providing a natural framework for continuous control. Actor–critic algorithms \cite{konda1999actor} unify value-based and policy-based paradigms by coupling a learned value function with a parameterized policy, improving variance properties and sample efficiency. Within this framework, deterministic policy gradient methods \cite{silver2014deterministic} exploit the structure of continuous action spaces by replacing stochastic policies with deterministic ones, achieving a more efficient gradient estimate. Deep Deterministic Policy Gradient (DDPG) \cite{ddpg} extends this idea using deep neural network approximators and off-policy learning, making it particularly suitable for high-dimensional, continuous control problems such as curling, where precise actuation is essential due to the nature of the game (see \cite{sumiea2024deep} for a recent systematic review of DDPG-based methods).

The characteristics of the game of curling outlined above, combined with the demands they place on sequential decision-making under uncertainty, make it a compelling and rigorous benchmark for reinforcement learning research. In this work, we formalize curling within the classical RL framework and demonstrate that, with careful problem design, an actor-critic method can learn tactically meaningful policies from simulation alone, without recourse to human gameplay data. In particular, we leverage the finite horizon nature of the game and  apply backward induction to reduce the computational complexity. We then solve each stage of the DDPG \cite{ddpg} algorithm to learn both a policy (actor) and a Q-function (critic); this stage-wise decomposition is related to, though distinct from, other recent adaptations of DDPG to episodic and finite-horizon settings, e.g., via terminal-value regularization \cite{guo2024deep} or dedicated exploration and replay strategies for sparse, terminal rewards \cite{futuhi2024etgl}. By training the last stage first, we use it to evaluate the resulting state of the previous stage without simulating all future stages. To generate samples to learn from, we use an adapted version of the Python-based simulation of curling developed by Michael Brunner \cite{brunner}.

A distinctive feature of the resulting framework is that the learned critic assigns a value to every action available in a given rink configuration, which is precisely the primitive required to \emph{compare} tactical alternatives rather than merely to select one. Building on this capability, we envision the proposed module being deployed in the three operational modes illustrated in Figure~\ref{fig:athlete-in-the-loop}. We stress that these describe intended applications of the framework, and delineate the scope of the tool we aim to build, rather than functionality evaluated in the present paper, whose experimental scope is confined to simulation:

\begin{itemize}
    \item \textbf{Tactical training} -- The athlete competes against an AI-controlled opponent, alternately selecting tactical actions. The physical execution of the actions is simulated and is not coupled to human motor performance. This mode is aimed at developing tactical awareness across a variety of game situations. The AI is used both to suggest actions to the athlete and to generate the opponent’s responses.
    
    \item \textbf{Athlete-in-the-loop training} -- On a real rink, the athlete selects and physically executes tactical actions, thereby jointly training tactical decision-making and technical execution. The outcome of each action is acquired through sensing. The AI simulates the opposing team’s response, which is then physically realized on the rink (e.g., by repositioning rocks), thus closing the feedback loop between AI decision-making and physical execution. See the upper panel of Figure~\ref{fig:athlete-in-the-loop}.
    
    \item \textbf{In-game decision support} -- The AI system observes the current state of the rink and provides recommended actions (or a ranked set of candidate actions) to the athlete. The athlete executes the selected action, after which the opposing team responds with its own move. See the lower panel of Figure~\ref{fig:athlete-in-the-loop}.
\end{itemize}

Interestingly, curling remains comparatively underexplored from a tactical perspective.  Most existing studies adopt either a statistical approach, focusing on shot outcomes or player tendencies in isolation, or a strategic perspective, analyzing decisions at an end-to-end level. 
In contrast, relatively few models address the tactical decision-making processes that govern the implementation of higher-level strategies. 
Furthermore, recent rule changes introduced in 2018 and 2023—most notably the five-rock free guard zone—have rendered much of the earlier literature  obsolete, thereby reinforcing the need for updated and systematic analyses.
Most curling-related literature focuses on statistical models for end-to-end strategy, particularly regarding whether blanking or scoring a single point in certain ends is preferred \cite{curling:markov, curling:markov2}.
Closer to our approach, \cite{rl-in-curling} propose a deep RL framework that combines kernel-regression Monte Carlo tree search with self-play reinforcement learning to handle the continuous action space of simulated curling, winning an international digital curling competition; unlike our work, however, their method targets individual shot execution rather than full tactical planning over an entire end, and does not exploit the finite-horizon structure of the game.
In \cite{wyss}, Wyss analyzed openings (the first five rocks of an end) and how they are influenced by the score difference. In \cite{brunner}, different approaches for tactical decision making are developed and compared.
Similarly, \cite{ahmad2016action} address the related problem of hammer-shot selection under a continuous, stochastic action space via Delaunay triangulation, reporting performance exceeding that of Olympic-level curlers, though again restricted to isolated shot decisions rather than full end-to-end tactical planning.
Most of these approaches rely, however, on supervised learning techniques using expert evaluations to train, with \cite{rl-in-curling} as a notable exception, relying instead on self-play reinforcement learning.
Additionally, the continuous action space of curling was simplified to a softmax output layer consisting of 305 possible calls.
As pointed out in \cite{brunner}, this discretization is not sufficient when compared to a continuous decision-making entity, such as human players.
Especially for a large number of rocks, there can be more than 305 possible calls.
Restricting the action space to a fixed set of predefined calls introduces bias and may prevent the agent from discovering truly optimal actions.
We aim to overcome this limitation in this work.

A concurrent work by Son et al.~\cite{son2026training} independently proposes a unified reinforcement learning framework for curling, formulating a single end as an episodic MDP and training a policy via Soft Actor-Critic within a two-phase curriculum, using a self-play mechanism to train entirely within a custom simulator; the resulting agent's shot patterns are subsequently compared against real elite match records to assess their alignment with human play. Our work differs primarily in the underlying algorithm: we adopt a deterministic actor-critic method (DDPG) explicitly restructured to exploit the finite-horizon nature of curling via backward induction, solving each stage of the game separately rather than optimizing the policy end-to-end.

After formulating the problem in Section \ref{sec:problem_formulation}, we introduce our RL methodology in detail in Section \ref{sec:methodology}. 
In Section~\ref{sec:results}, we show that the critic successfully approximates the general shape of the Q-function, providing a solid foundation for training effective policies. Our proposed method is then evaluated on a simplified game variant against a heuristic baseline that encodes expert curling knowledge, and is shown to match its performance despite being trained without any domain knowledge or human-annotated data, demonstrating that meaningful curling tactics can be acquired in a fully self-supervised manner. Finally, Section~\ref{sec:conc} concludes the paper with a discussion of the key takeaways and an outlook on promising directions for future work.

\section{Problem formulation and preliminaries}
\label{sec:problem_formulation}
\subsection{Game-theoretic modeling of Curling}
We formulate the game of curling as a two-player turn-based stochastic game, allowing us to leverage the advancements in reinforcement learning and game theory to derive the optimal or close-to-optimal policy in this setting. The stochasticity stems from the imperfect execution of actions. In this section we define the elements of this game and motivate their choice.

The game is abstracted on a turn-by-turn basis; that is, we measure the state of the game only after all rocks have come to rest, thus ignoring the intra-shot sliding dynamics. 

The state of the game $s_k \in \mathcal{S}$ at stage $k$ is defined as:
\begin{align*}
    s_k = (d, e, h, k, \Vec{p}),
\end{align*}
where:
\begin{itemize}
    \item [$d$] $\in \mathbb{Z}$ is the score difference between the two teams. Since only the relative score matters within a tournament or game, the absolute score is irrelevant.
    \item [$e$] $\in \mathbb{N}$ is the current end of the game.
    \item [$h$] $\in \{0,1\}$ denotes the team holding the hammer in the ongoing end. Conversely, $\bar{h}=1-h$ denotes the team going first.
    \item [$k$] $\in \{0,\hdots, N\}$ is the index of the next shot in the end, where $k=N$ corresponds to the terminal state at which the score is evaluated. In classical curling, $N=16$, while other variants exist, such as mixed doubles with $N=10$.
    \item [$\Vec{p}$] is the set of rocks currently in play, with $|\Vec{p}| \leq N$. Each element is a triple $(t, x, y)$, where $t \in \{0, 1\}$ is a team indicator and $(x, y) \in \mathbb{R}^2$ are the spatial coordinates of the rock.
\end{itemize}

The state is assumed to be fully observable at all times, including in situations where it is not discernible by eye whether a rock touches the house or centerline, which determines its protection status under the free guard and no-tick zones. In such cases, if the teams cannot agree, the rock's status is determined by an umpire.

\begin{figure*}[tb]
    \centering
    \includegraphics[scale=0.45]{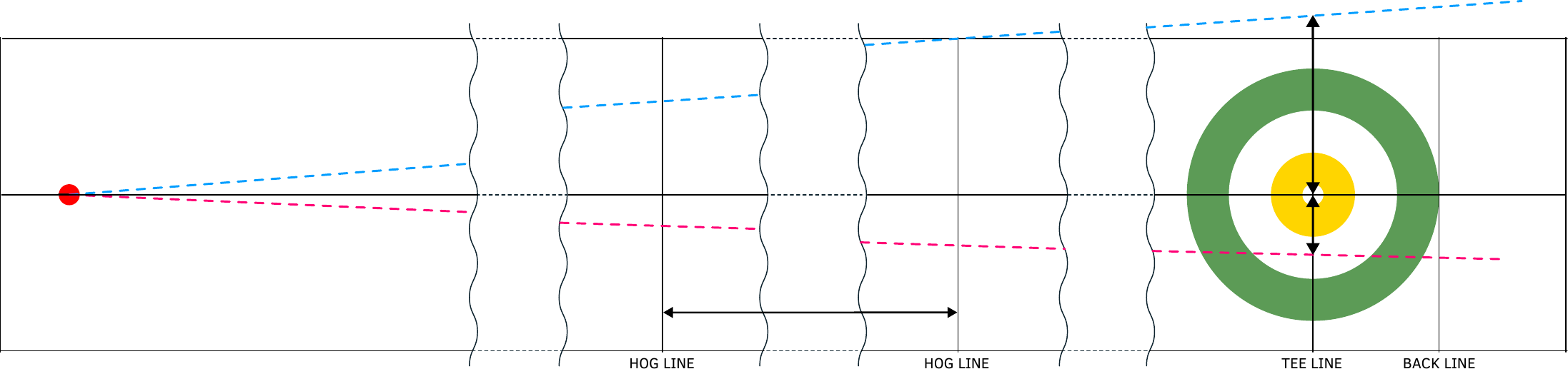}
    \put(-230,20){$\tau$}
    \put(-72,41){$\xi$}
    \put(-72,76){$\xi_{max}$}
    \caption{Visualization of the action space of the game: $\tau$ is the hot-to-hog time; $\xi \in [-\xi_{\max}, \xi_{\max}]$ is the coordinate of the aiming point.}
    \label{fig:action-space}
\end{figure*}

The action taken by a player at stage $k$ is denoted as $a_k \in \mathcal{A}$ and defined as the tuple:
\begin{align*}
    a_k &= (\tau, \xi, \omega) \in \mathcal{A} \subseteq 
    \mathbb{R}_{[t_1,t_2]} \times \mathbb{R}_{[-\xi_{\max},\xi_{\max}]} \times 
    \{\texttt{cw}, \texttt{ccw}\},
\end{align*}
where:
\begin{itemize}
    \item [$\tau$] $\in [t_1, t_2]$ is the time it takes the rock to travel between the two hoglines, which is the standard way to measure rock speed in curling as it is unaffected by sweeping and robust to measurement imprecision\footnote{Players are allowed to use a stopwatch on the ice for this purpose.}. Here, $t_1$ is the minimum time corresponding to the fastest physically deliverable rock, and $t_2$ is the maximum time such that the rock still crosses the far hogline and remains in play.
    \item [$\xi$] $\in [-\xi_{\max}, \xi_{\max}]$ is the $x$-coordinate of the aiming point, fixed at the teeline for consistency. The quantity $\xi_{\max}$ denotes the maximum meaningful lateral offset at which a rock can be played, corresponding to the line from the hack passing through the corner of the playing area, as illustrated in Figure~\ref{fig:action-space}.
    \item [$\omega$] $\in \{\texttt{cw}, \texttt{ccw}\}$ is the direction of the rock's rotation about the vertical axis (clockwise or counterclockwise). This rotation causes the rock to deviate laterally from a straight path as it travels, an effect known as \textit{curl}, from which the sport takes its name. The magnitude of the angular velocity is known to have negligible influence on the trajectory and is therefore not actively varied by players; only its sign is relevant.
\end{itemize}

The action selected by the policy will be referred to as a \textit{call}, representing the intended action prior to any noise. In practice, the executed action deviates from the call due to player skill and rink conditions. To model this imprecision, we add zero-mean Gaussian noise to the initial speed and lateral direction of the rock. The corresponding standard deviations $\sigma_\tau$ and $\sigma_\xi(a_{k})$ are modeled as a Gaussian Mixture Model following~\cite{brunner}, capturing the variable precision observed even among top-level players. The resulting noise at stage $k$ is denoted as:
\begin{equation*}
    w_{k} \sim \mathcal{N}\bigl(0,\, \Sigma(a_{k})\bigr), \qquad
    \Sigma(a_{k}) = \operatorname{diag}\bigl(\sigma_\tau^{2},\, \sigma_\xi^{2}(a_{k})\bigr),
\end{equation*}
with all samples drawn independently across stages.
The chosen noise model corresponds to the precision of an average high-level player.

Notice that we have not modeled \emph{sweeping}. 
This is barely restrictive: sweeping is mainly an aid to correct small imprecisions in the initial speed and direction and to compensate for ice conditions. 
Therefore, sweeping can be accounted for in the overall precision of a player. 

As is standard in finite-horizon games~\cite{silver2017mastering}, the intermediate stage rewards are set to zero:
\begin{equation}
    R(s_{k}, a_{k}) = 0, \qquad \forall\, k < N,
\end{equation}
while the terminal reward is defined as:
\begin{equation}
   R(s_{N}) = \textit{score}(s_{N}), \label{eq:last-stage}
\end{equation}
where $\textit{score}(\cdot)$ evaluates the final rock configuration once all $N$ rocks have been delivered. 
By symmetry of the game, $\textit{score}(s_N)$ returns a positive value if the hammer team scores, and a negative value if the non-hammer team scores.

Having defined the states, actions, and rewards in the context of curling, we are ready to formulate the policy optimization problem. The goal of team $h$ is to find a policy $\mu_h(\cdot)$ that maximizes the expected cumulative reward, while team $\bar{h}$ seeks a policy $\mu_{\bar{h}}(\cdot)$ that minimizes the same quantity. Defining the state transition kernel as:
\begin{equation*}
    P(s_{k+1} \mid s_{k},\, a_{k},\, w_{k}(a_{k})),
\end{equation*}
we can formulate the problem as follows.
Find:
\begin{align*}
    %&\text{Find:} \nonumber\\
    &\mu_h^* = \arg\max_{\mu_h} \min_{\mu_{\bar{h}}} \; 
    \mathbb{E}_{w_{k}}\left[\sum_{k=0}^{N} R(s_k, a_k) \,\bigg|\, s_0,\, \mu_h,\, 
    \mu_{\bar{h}}\right] \\[6pt]
    &\mu_{\bar{h}}^* = \arg\min_{\mu_{\bar{h}}} \max_{\mu_{h}} \; 
    \mathbb{E}_{w_{k}}\left[\sum_{k=0}^{N} R(s_k, a_k) \,\bigg|\, s_0,\, \mu_h,\, 
    \mu_{\bar{h}}\right] 
\end{align*}
subject to:
\begin{align*}
    &s_{k+1} \sim P(s_{k+1} \mid s_{k},\, a_{k},\, w_{k}(a_{k})), 
    \ \  k = 0, \ldots, N\!-\!1 \\[4pt]
    &a_{k} = \begin{cases} \mu_h(s_{k}) & \text{if } k \in \mathcal{T}_h, \\ 
    \mu_{\bar{h}}(s_{k}) & \text{if } k \in \mathcal{T}_{\bar{h}}, \\
    \emptyset & \text{if } k = N,\end{cases}
\end{align*}
where the expectation is taken over the stochastic state transitions induced by the action noise $w_{k}(a_{k})$, $N$ is the total number of throws per end, and $\mathcal{T}_h$, $\mathcal{T}_{\bar{h}}$ denote the index sets of throwing turns belonging to team $h$ and team $\bar{h}$, respectively. This formulation casts the problem as a two-player zero-sum stochastic game over a finite horizon. Due to the continuous state and action spaces and the sensitivity of state transitions to small perturbations in the executed action, an exact solution is intractable, and we therefore resort to deep reinforcement learning to approximate the optimal policies.

\subsection{Reinforcement Learning Concepts}
Reinforcement learning (RL) \cite{sutton1998reinforcement} provides a framework for learning decision-making strategies through interaction, and extends naturally to competitive multi-agent settings. As formalized above, our problem is a two-player turn-based stochastic game, described by the tuple $(\mathcal{S}, \mathcal{A}, P, r_1, r_2)$, in which only one player acts at each step $k$ according to its policy $\pi_i$, while the other observes the state, and the rewards satisfy $r_1 = -r_2$. Both agents are trained simultaneously via \emph{self-play} \cite{chakraborty2014multiagent}: each player learns its policy by interacting with copies of itself, allowing strategies to co-evolve and adapt to the opponent's behavior. For continuous action spaces, Deep Deterministic Policy Gradient (DDPG) \cite{ddpg} employs an actor--critic architecture \cite{ac}, where each agent learns a deterministic policy $\mu_{\theta_i} : \mathcal{S} \rightarrow \mathcal{A}$ together with a critic estimating $Q_i^{\mu}(s, a)$, and where experience replay and target networks are used to stabilize training.

\section{Leveraging Backward Induction and Deep RL for Curling Strategy Learning}
\label{sec:methodology}
\subsection{Backward Induction}

Since $N < \infty$ and, more specifically, fixed and small, we can exploit the finite-horizon structure of the game via backward induction. At the terminal state $s_N$ (i.e., after the last rock has been played and all rocks have come to rest), no further actions are possible, i.e., $a_N=\emptyset$, and the action value function reduces to:
\begin{equation}\label{eq:v-last-stage}
    Q(s_N,a_N) = R(s_N),
\end{equation}
where $R(s_N)$ is the terminal reward defined in Equation~\eqref{eq:last-stage}. Moving one step backward, the action-value function at stage $N-1$ is given by the Bellman operator~\cite{bellman1957markovian}:
\begin{equation}
\label{eq:bellman}
    Q(s_{N-1}, a_{N-1}) = \underbrace{R(s_{N-1}, a_{N-1})}_{=\,0} + 
    \mathbb{E}_{w_{N-1}}\big[Q(s_N,a_N)\big].
\end{equation}
Since no intermediate reward is assigned when playing a rock, the first term vanishes and Equation~\eqref{eq:bellman} simplifies to:
\begin{equation}
    Q(s_{N-1}, a_{N-1}) = \mathbb{E}_{w_{N-1}}\big[Q(s_N,a_N)\big].
\end{equation}
The optimal policy for the last rock is then obtained as:
\begin{equation}\label{eq:mu-last-stage}
    \mu(s_{N-1}) = \underset{a \in \mathcal{A}}{\arg\max}\ Q(s_{N-1}, a).
\end{equation}
Having solved the last stage, we can proceed backward to stage $N-2$. The corresponding action-value function is obtained by substituting the already-computed optimal policy at the previous stage:
\begin{equation}
    Q(s_{N-2}, a_{N-2}) = \mathbb{E}_{w_{N-2}}\big[Q\big(s_{N-1},\, 
    \mu(s_{N-1})\big)\big],
\end{equation}
and the optimal policy at stage $N-2$ follows analogously:
\begin{equation}
    \mu(s_{N-2}) = \underset{a \in \mathcal{A}}{\arg\max}\ Q(s_{N-2}, a).
\end{equation}
This procedure generalizes recursively to all stages $k \in \{0, \ldots, N-1\}$:
\begin{equation}\label{eq:general}
    \begin{aligned}
        Q(s_k, a_k) &= \mathbb{E}_{w_k}\big[Q\big(s_{k+1},\, \mu(s_{k+1})\big)\big], \\[4pt]
        \mu(s_k) &= \underset{a \in \mathcal{A}}{\arg\max}\ Q(s_k, a),
    \end{aligned}
\end{equation}
where $s_{k+1} \sim P(s_{k+1} \mid s_k, a_k, w_k(a_k))$, and $Q(s_N,a_N) = R(s_N)$. We note that neither of these equations admits a closed-form solution, as the state transition distribution captures both complex rock collision dynamics and stochastic action noise. Furthermore, the action-value function $Q$ is not restricted to any specific functional form (e.g., linear or convex), which renders the $\arg\max$ operation in Equation~\eqref{eq:general} analytically intractable in general. Consequently, we adopt a deep reinforcement learning approach to approximate both $Q$ and $\mu$ at each stage through repeated interaction with the environment.

\subsection{Training with Reinforcement Learning}
As mentioned above, we adopt a deep reinforcement learning approach to tackle the optimization problem in Equation \eqref{eq:general}.
Specifically, we implement the state-of-the-art DDPG algorithm, adapted for our two-player zero-sum game, adopting \textit{self-play} \cite{chakraborty2014multiagent}, and training a separate agent for each stage.

DDPG is particularly well suited in our setting for several reasons. First, it naturally accommodates continuous action spaces and benefits from the dense learning signal provided by the critic, enabling stable and efficient policy updates. Second, given its model-free nature, it eliminates the need to explicitly model or invert the state transition dynamics—an intractable task in our environment due to its complexity and stochasticity.
Additionally, it is worth motivating this choice explicitly against the stochastic actor-critic methods that are more commonly adopted in continuous control, and that have recently been applied to curling in concurrent work \cite{son2026training}. Algorithms such as Soft Actor-Critic maximize an entropy-regularized objective, deliberately maintaining a stochastic policy in order to improve exploration and robustness. In our setting this is a mismatch on two counts. First, the environment already injects substantial stochasticity through the execution noise $w_k$, so that the mapping from a call to an outcome is far from deterministic; additional policy entropy is therefore largely redundant, and it biases the learned policy away from the single action that an athlete should in fact be advised to attempt. Second, and more fundamentally, backward induction requires evaluating each stage at the specific action prescribed by the already-solved subsequent stage, as in Equation~\eqref{eq:general}. A deterministic policy $\mu(s_{k+1})$ makes this substitution exact, whereas an entropy-regularized policy would require marginalizing over the action distribution at every stage, reintroducing precisely the nested expectations that the backward decomposition is designed to eliminate. A similar argument applies to twin-critic variants such as TD3: the terminal-only reward structure bounds the return within $[-N/2, N/2]$, which limits the scope for the value overestimation that those methods are designed to correct.

\begin{figure}
    \centering
    \includegraphics[width=\linewidth]{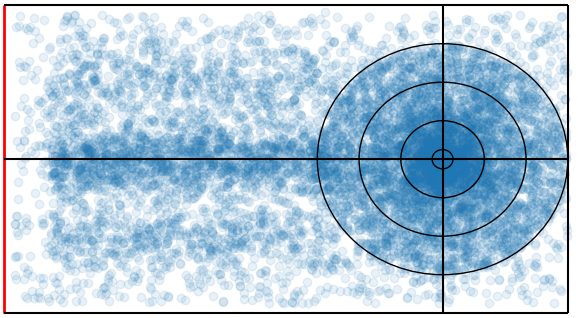}
    \caption{Distribution of the rocks' starting positions. The rocks have a $55\%$ chance to be in the house, $15\%$ to be a center guard, $10\%$ to be a corner guard, and a $20\%$ chance to be placed uniformly in the playing area.}
    \label{fig:rock-dist}
\end{figure}

The overall training process is split into two phases: \textit{pre-training} and \textit{fine-tuning}.
During pre-training, we randomly sample at each iteration a state of the game\footnote{Positions with a randomly selected number of rocks for each team within the maximum possible number at the current stage.} with a suitable generation strategy. Specifically, the rock positions are selected according to the distribution depicted in Figure \ref{fig:rock-dist}, which has been inspired from \cite{brunner}, and adapted here to create more realistic scenarios for training, by mainly favoring positions in the house, close to the button, and center guards, preventing easy access to rocks close to the button.

After this initial pre-training phase, the reinforcement learning agents are fine-tuned on more realistic game scenarios, generated by the policies of the earlier stages. 
In particular, we generate random positions for the second-to-last stage, then take an action according to the corresponding policy, and use the resulting position to train the last stage.
After several samples are generated in this way, and the agents have been trained on the last two stages, we move on to stage $N-3$, and the process continues iteratively: random positions for this stage are generated and the game is played until the \textit{end} is over.
The main advantage of this phase is that the networks are exposed to situations that are more realistic than those produced by purely random generation, as they emerge from previously learned policies. Moreover, gradually extending the game horizon improves the performance of later stages, whose learned policies and value estimates are then used to train earlier stages. The full-game training (\textit{fine-tuning}) is summarized in Algorithm \ref{algo}.

\begin{algorithm}
    \caption{Full-Game Training}\label{algo}
    \begin{algorithmic}
    \Require
    \State $actor_k, critic_k, \leftarrow$ Pre-trained actor and critic networks for each stage $k$
    \State $\mathcal{B}_k \leftarrow$ Empty memory buffer for each stage $k$
    \State $\{M_k\}_{k=0,...,N-2} \leftarrow$ \# of training iterations for each stage
    \State $\sigma_{expl}^2 \leftarrow$ Exploration noise variance
    \State
    \For{start stage $k_0 = N-2 \textbf{ to } 0$}
    \State $i=0$
    \While{$i\leq M_{k_0}$}
    \State $j \leftarrow k_0$
    \State $\state{j} \leftarrow$ random position for stage $k_0$
    \Repeat
    \State $\action{j} \leftarrow actor_j(\state{j}) + \mathcal{N}(0, \sigma_{expl}^2)$
    \State Take action $\action{j}$ in the environment
    \State Move to $\state{j+1}\sim P$ and observe $r_j$
    \State Add $(\state{j}, \action{j}, \state{j+1}, r_j)$ to $\mathcal{B}_j$
    \State $j = j+1$
    \State $i = i+1$
    \Until{$j=N$}
    \For{stage $k' = k_0 \textbf{ to } N-1$}
    \State Update $actor_{k'},critic_{k'}$ using buffer $\mathcal{B}_{k'}$
    \State and the DDPG algorithm \cite{ddpg}
    \EndFor
    \EndWhile
    \EndFor
    \end{algorithmic}
\end{algorithm}

\section{Experimental results}
\label{sec:results}
This section presents the experimental evaluation of the proposed approach. It first describes the simulation setup used to conduct the experiments, then introduces the baseline method considered for comparison, and finally reports and discusses the obtained results.

\subsection{Simulation setup}
\label{subsec:sim-setup}
The results presented in this section have been obtained by considering a reduced version of the curling game, mainly to lighten the computational load required.
In particular, we have considered a 4-rock variant of the game, meaning that each team plays two rocks in a single end.
The agents were trained on 1 million of transitions for each stage, including the fine-tuning phase, where all stages were trained together. For our experiments, we employed the simulator developed in \cite{brunner}.

Both the actor and the critic are parameterized as neural networks with two hidden layers of equal width and ReLU activations. The proposed architecture assigns a dedicated DDPG agent to each game stage, with a hidden layer size of 256 neurons. To assess the impact of model capacity and architectural choices, we conduct an ablation study considering two additional hidden layer sizes — 64 and 128 neurons — as well as a simplified variant in which a single DDPG agent is shared across all stages, serving as a lightweight alternative to the full per-stage architecture. All configurations are then benchmarked against an \textit{expert heuristic} baseline, formally defined in Algorithm~\ref{alg:greedy}. The baseline selects at each stage the action that maximizes the immediate scoring outcome, without any lookahead into future stages, and is therefore myopic by construction. While this makes it clearly suboptimal in the general case, it is a deliberately demanding point of comparison in our simplified setting, where each team plays only two rocks per end. With so few shots available, the margin for long-term tactical planning is limited, and a well-executed immediate scoring strategy already captures most of the available tactical depth. The baseline should therefore not be understood as a naive competitor, but as a compact encoding of expert curling knowledge that is close to optimal at this horizon — a point we quantify below.

\begin{algorithm}
\caption{Expert heuristic baseline} \label{alg:greedy}
\begin{algorithmic}
\If{own rock is in the best position}
\State Draw as close to it as possible
\ElsIf{opponent's rock is in the best position}
\State Hit that rock
\Else \Comment{No rock in the house}
\State Draw to the button
\EndIf
\end{algorithmic}
\end{algorithm}

Before evaluating the learned policies, it is instructive to quantify how much room for improvement this simplified variant actually leaves. To this end, we let the heuristic baseline play against itself over 100 simulated ends: the team holding the hammer won 97\% of them. In the 4-rock game, the last-rock advantage is therefore close to decisive on its own, and win rates of approximately 97\% with hammer and 3\% without constitute an effective ceiling against which any policy must be measured. We report this reference alongside our results in Table~\ref{tab:results}, and we return to its implications when discussing them.

\subsection{Results and discussion}
Performance is assessed through two complementary approaches. First, the learned Q-function and policy are analyzed using two representative scenarios of curling games; this is where the added value of the proposed method is most apparent, as the critic provides a dense evaluation of the entire action space that a rule-based policy cannot offer. Second, we quantitatively compare our proposed method against the heuristic baseline across many simulated games, interpreting the outcome in light of the hammer-advantage reference quantified in Section~\ref{subsec:sim-setup}.
\subsubsection{Scenario evaluation 1} \label{subsubsec:scenario1}

\begin{figure}
    \centering    \includegraphics[scale=0.45]{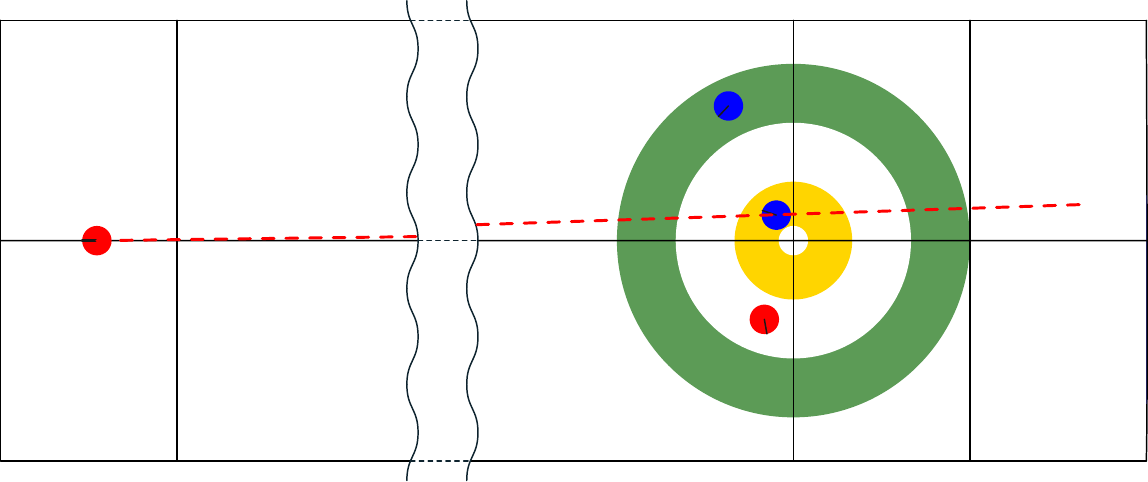}
    \caption{Learned actor policy for the 4-rock minigame, in a configuration offering an unobstructed path to the house. The game is tied in the last end, with team red delivering the last rock and the blue shot rock lying at the center of the house. The dashed red line indicates the trajectory of the action selected by the policy, a direct takeout on the shot rock.}
    \label{fig:a1}
\end{figure}
    
\begin{figure}
    \centering        \includegraphics[width=\linewidth]{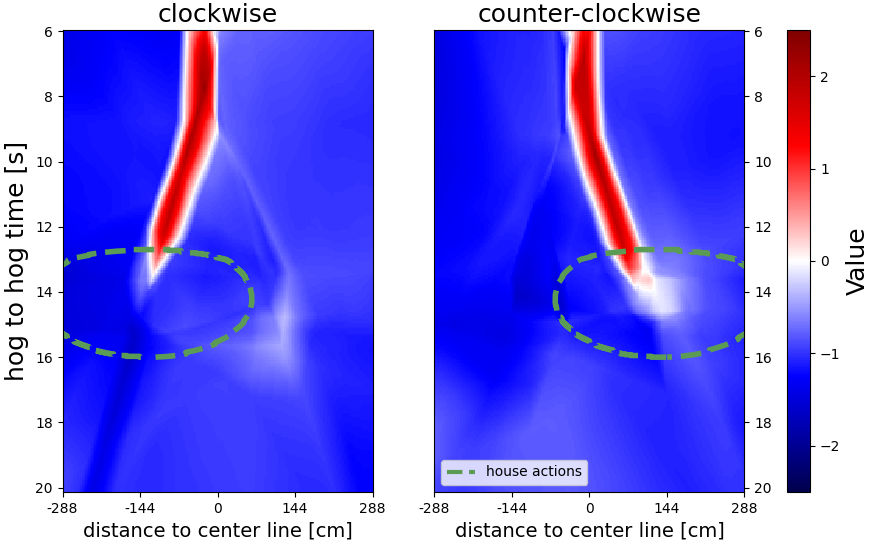}
    \caption{Critic's Q-function evaluation for the game state in Figure \ref{fig:a1}. Each pixel represents the expected score for the corresponding action. The two subplots correspond to clockwise (left) and counter-clockwise (right) handle orientations. Red regions indicate takeout actions on the blue shot rock at house center, while color intensity reflects expected score magnitude.
    }
    \label{fig:c1}
\end{figure}

Figure \ref{fig:c1} presents the critic's evaluation of the game state shown in Figure \ref{fig:a1}, where the final rock of the game is about to be played by team red.
The action space is discretized into pixels, with each pixel's value representing the corresponding Q-function estimate. The color scale indicates expected scoring outcomes: red (blue) denotes actions favoring team red (blue), with color intensity reflecting the magnitude of the expected score.

The critic visualization comprises two subplots, corresponding to the binary handle parameter $\omega\in\{\texttt{cw}, \texttt{ccw}\}$. The left subplot displays Q-values for clockwise handles, which induce rightward trajectory deviation (relative to the velocity vector). The right subplot shows counter-clockwise handle values, which produce leftward deviation.

The horizontal axis represents the directional component of the action, while the vertical axis denotes hog-to-hog time.
Hog-to-hog time spans the interval $[6, 20.1]$ seconds, where $6s$ represents the physical limit for elite player delivery speed and $20.1s$ denotes the minimum speed required for a rock to remain in play after crossing the hogline. Consequently, actions positioned higher in the plot correspond to faster deliveries.
The dashed green contour delineates actions that place the rock within or touching the house boundary (the minimum requirement for scoring eligibility), assuming perfect execution without noise. The dashed red trajectory in Figure \ref{fig:a1} shows the intended path for a rock delivered at the policy's selected action, though action noise applied during execution may cause deviation from this nominal trajectory.

Moreover, the two red regions in Figure \ref{fig:c1} correspond to takeout actions on the shot rock. Their asymmetric shape reflects the physics of rock trajectories: lateral deviation decreases with delivery speed, requiring the aiming point to approach the target rock more directly. This effect diminishes at higher speeds as lateral deviation approaches zero, evident in the nearly vertical shape of the red region between $8s$ and $6s$.
The gradual transition from red to blue values reflects the decreasing probability of successful contact with the blue rock. Action noise acts as a smoothing factor on the Q-function, which the critic's network architecture captures effectively.
For counter-clockwise actions, a white region appears at the house center, representing actions that may either position the shooter closer to the current shot rock or execute a soft collision that displaces it sufficiently for the shooter to become the new best rock. However, action noise prevents consistent execution of this outcome, yielding an expected value near zero.

\begin{figure}
    \centering
    \includegraphics[width=\linewidth]{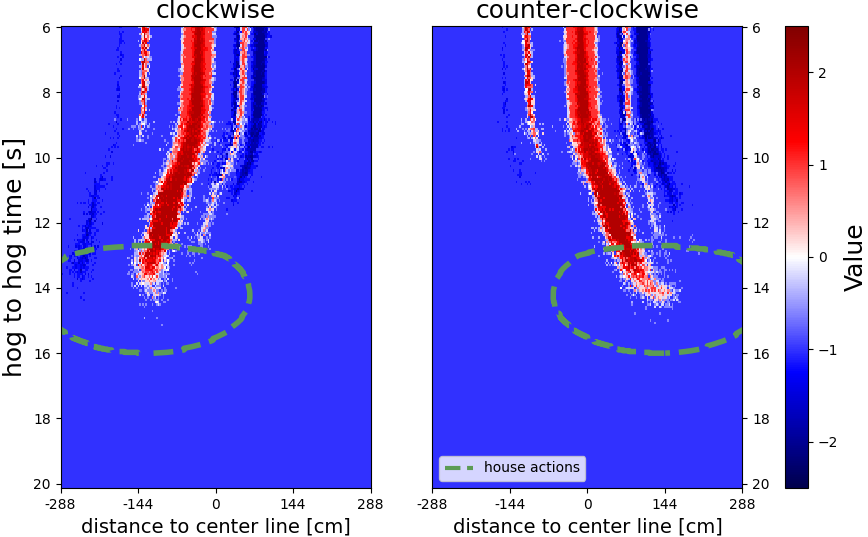}
     \caption{Monte Carlo reference estimate of the Q-function for the game state in Figure \ref{fig:a1}, obtained by discretizing the action space and averaging the outcomes of five simulated executions per action. The estimate provides a reference for the \emph{structure} of the Q-function rather than a pointwise ground truth.}
    \label{fig:true-c1}
\end{figure}
Finally, Figure \ref{fig:true-c1} presents a Monte Carlo reference estimate of the Q-function for the state in Figure \ref{fig:a1}, constructed by discretizing the action space and simulating each action five times, with the mean outcome score assigned to each cell. We deliberately refrain from calling this a ground truth: five executions per cell are far too few to resolve the expectation over the action noise, and the speckled texture visible in Figure \ref{fig:true-c1} is therefore largely sampling variance rather than genuine structure of $Q$. The comparison with Figure \ref{fig:c1} should accordingly be read qualitatively, in terms of which regions of the action space are identified as valuable, and not as a pointwise accuracy assessment.

Read in this way, the agreement is good: the learned critic recovers the position, curvature, and asymmetry of the primary high-value band, which is expected given that takeout actions on the shot rock receive extensive exploration during training, as they represent strong strategic options across many game states. It is worth noting that the smoothness of the learned critic is not merely an artifact of function approximation, but also acts as a form of variance reduction: by generalizing across neighbouring actions, the network aggregates information from far more than the five samples available to each cell of the reference, so that in sparsely sampled regions its estimate of the expected score may in fact be closer to the true $Q$ than the Monte Carlo reference itself.

Nevertheless, some regions adjacent to the main high-value band remain inadequately learned. For instance, the smaller high-value region to the left of the primary area corresponds to double-takeout actions removing both blue rocks. While this action offers no advantage in the current configuration, it could prove superior in alternative scenarios---for example, when a guard protects the blue shot rock from direct contact, or when the red rock lies farther from center than the second blue rock.

\subsubsection{Scenario evaluation 2}

\begin{figure}
    \centering
    \includegraphics[scale=0.45]{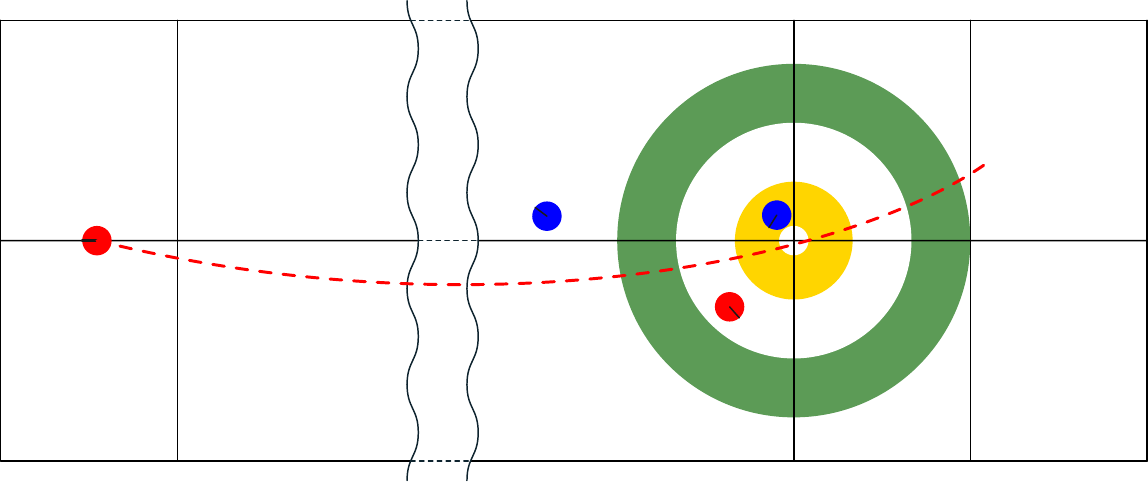}
    \caption{Learned actor policy for the same scenario as Figure~\ref{fig:a1}, with the sole difference that the second blue rock now sits as a guard blocking the direct path to the shot rock. The dashed red line indicates the trajectory of the action selected by the policy: a markedly slower delivery that curls around the guard.}
    \label{fig:guard}
\end{figure}

\begin{figure}
    \centering
    \includegraphics[width=\linewidth]{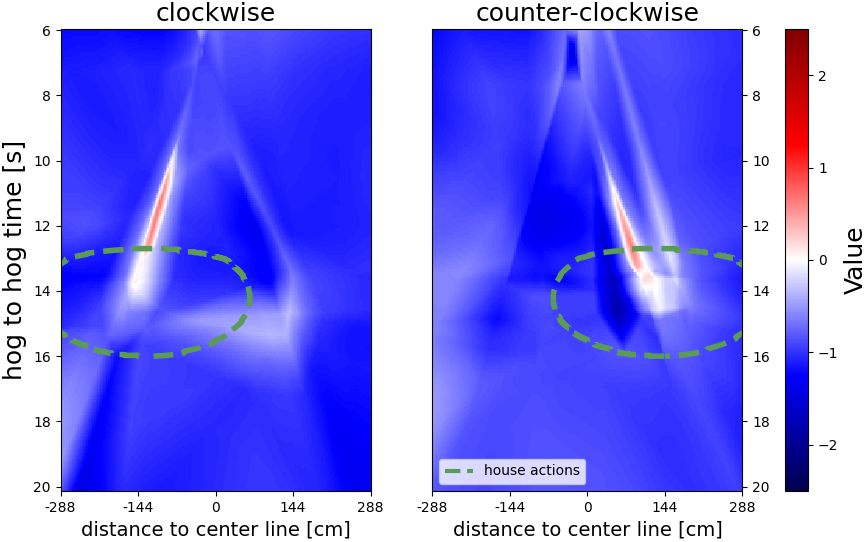}
    \caption{Critic's Q-function evaluation for the game state in Figure \ref{fig:guard}. Each pixel represents the expected score for the corresponding action. The two subplots correspond to clockwise (left) and counter-clockwise (right) handle orientations. Red regions indicate takeout actions on the blue shot rock at house center, while color intensity reflects expected score magnitude.
    }
    \label{fig:critic-guard}
\end{figure}

Figure \ref{fig:guard} presents a similar scenario with respect to Figure \ref{fig:a1}, with one critical modification: the blue rock previously in the green ring now obstructs the policy's intended trajectory.
The critic evaluation (Figure \ref{fig:critic-guard}) shows the red takeout region has largely disappeared due to the blocked direct path. Only a small red area remains, containing the actor's selected action. We can notice that the policy now selects a significantly slower delivery than previously.
This reduced speed is necessary because the guard blocks the direct trajectory. The rock must circumvent the guard, requiring lateral displacement after passing it -- a characteristic achievable only through slow deliveries, which exhibit increased sideways motion in the latter portion of their trajectory. Given the action noise applied to the intended shot, successful contact with the center blue rock remains probable across multiple executions.

The optimal outcome would displace the blue rock farther from center than the existing red rock, while positioning the shooter closer to the displaced blue rock's new location, yielding a two-point score. Even if one objective fails, the strategy maintains reasonable probability of scoring one point.

\subsubsection{Comparison against baseline}\label{subsubsec:tournament}
Table~\ref{tab:results} summarizes the performance of our best-performing \textit{Curling RL} agent, evaluated over 200 simulated ends against the heuristic baseline, 100 with hammer and 100 without. The results must be read against the hammer-advantage reference established in Section~\ref{subsec:sim-setup}. Holding the hammer, our agent wins 100\% of the ends, compared to the 97\% obtained by the heuristic itself in the same position; without the hammer, it steals 2\% of the ends, compared to the heuristic's 3\%. Neither difference is statistically significant at this sample size, so the two policies are indistinguishable in terms of win rate. We stress that this should not be read as a failure to improve upon the baseline, but rather as a property of the evaluation regime: at a horizon of four rocks, the hammer advantage alone already accounts for a 97\% win rate, leaving essentially no headroom for any policy to separate itself on this metric. What the comparison does establish is that a policy trained entirely through self-play, without any domain knowledge and without human-annotated data, attains the performance ceiling of a hand-crafted heuristic in which expert curling knowledge is encoded explicitly. 
That this ceiling is not attained trivially is confirmed by the remaining configurations — smaller hidden layer sizes (64 and 128 neurons) and the single shared-agent variant — which, trained on the same amount of data, remained consistently below both our agent and the heuristic reference. The win-rate metric thus saturates at the top of its range while still discriminating among weaker policies, which indicates that matching the reference is a non-trivial outcome rather than an artifact of the metric, and that both model capacity and a dedicated per-stage architecture contribute to achieving~it.

\begin{table}
     \caption{Win rates over 100 simulated ends per condition. The first row reports the hammer-advantage reference, obtained by letting the heuristic baseline play against itself; the second row reports our Curling RL agent against the same baseline.
     }
    \label{tab:results}
    \renewcommand{\arraystretch}{1.3}
    \begin{tabular}{lcc}
        \toprule
        & \textbf{With Hammer} & \textbf{Without Hammer} \\
        \midrule
        Heuristic vs. heuristic & 97\% & 3\% \\
        \textbf{Curling RL} vs. heuristic & \textbf{100\%} & 2\% \\
        \bottomrule
    \end{tabular}
\end{table}

\section{Conclusions and outlook}
\label{sec:conc}
We study the game of curling from a strategical decision-making point of view, and introduce a reinforcement learning-based AI module that can be utilized for tactical training, athlete-in-the-loop training and in-game decision support. By formulating the game as a finite-horizon, stochastic, and adversarial decision process with continuous state and action spaces, we demonstrate that tactical curling behavior can emerge through simulation alone, without explicit domain knowledge.

The method operates by iteratively training policies for each stage of the game, starting from the final rock and progressing backward. Earlier-stage actions are evaluated using value estimates from later stages. The learned agents consistently exhibit context-aware behavior, adapting their decisions to the current rock configuration and shot number. Particularly in short ends of four rocks per team, policies converged toward recognizable curling tactics, validating the potential of the approach.

Despite these advances, several limitations remain. The model considers only single ends and does not account for longer-term planning across multiple ends, which is central to high-level curling strategy. A related limitation concerns evaluation: in the 4-rock variant studied here, the hammer advantage alone determines the outcome of 97\% of the ends, so win rate saturates and cannot resolve differences between near-optimal policies. Assessing the benefit of long-horizon reasoning therefore requires longer ends, where the myopic nature of the heuristic baseline is expected to become a genuine handicap, or finer-grained metrics such as the expected number of points scored per end.
Nonetheless, we believe that this work provides a first step toward learning curling strategy through reinforcement learning and simulation-based self-play. It offers a potential tool for post-game analysis, on- and off-ice tactical training or strategy exploration, and establishes a methodological foundation for future research in curling and similar domains. Promising directions for extension include full-game tactical modeling,  improvement of the exploration techniques to better support athletes in evaluating risk-reward decisions, and adapting the algorithm to the opponent's skill level during training.

\printbibliography
\newpage
\vfill
\end{document}